\documentclass[acmtog,nonacm]{acmart}

\usepackage{booktabs} 
\usepackage{enumitem}
\usepackage[ruled]{algorithm2e} 
\usepackage[size=small]{caption}

\SetAlFnt{\small}
\SetAlCapFnt{\small}
\SetAlCapNameFnt{\small}
\SetAlCapHSkip{0pt}

\authorsaddresses{}  

\definecolor{lightblue}{RGB}{30,144,255} 

\begin{document}
\title{UltraZoom: Generating Gigapixel Images from Regular Photos}

\author{Jingwei Ma}
\affiliation{%
  \institution{University of Washington}
  \city{Seattle}
  \country{USA}
}

\author{Vivek Jayaram}
\affiliation{%
  \institution{University of Washington}
  \city{Seattle}
  \country{USA}
}

\author{Brian Curless}
\affiliation{%
  \institution{University of Washington}
  \city{Seattle}
  \country{USA}
}

\author{Ira Kemelmacher-Shlizerman}
\affiliation{%
  \institution{University of Washington}
  \city{Seattle}
  \country{USA}
}

\author{Steven M. Seitz}
\affiliation{%
  \institution{University of Washington}
  \city{Seattle}
  \country{USA}
}

\begin{abstract}
We present UltraZoom, a system for generating gigapixel-resolution images of objects from casually captured inputs, such as handheld phone photos.
Given a full-shot image (global, low-detail) and one or more close-ups (local, high-detail), UltraZoom upscales the full image to match the fine detail and scale of the close-up examples.
To achieve this, we construct a per-instance paired dataset from the close-ups and adapt a pretrained generative model to learn object-specific low-to-high resolution mappings.
At inference, we apply the model in a sliding window fashion over the full image.
Constructing these pairs is non-trivial: it requires registering the close-ups within the full image for scale estimation and degradation alignment.
We introduce a simple, robust method for getting registration on arbitrary materials in casual, in-the-wild captures. Together, these components form a system that enables seamless pan and zoom across the entire object, producing consistent, photorealistic gigapixel imagery from minimal input.
For full-resolution results and code, visit our project page at \textcolor{lightblue}{\href{https://ultra-zoom.github.io}{ultra-zoom.github.io}}.

\end{abstract}

\begin{teaserfigure}
    \includegraphics[width=\textwidth]{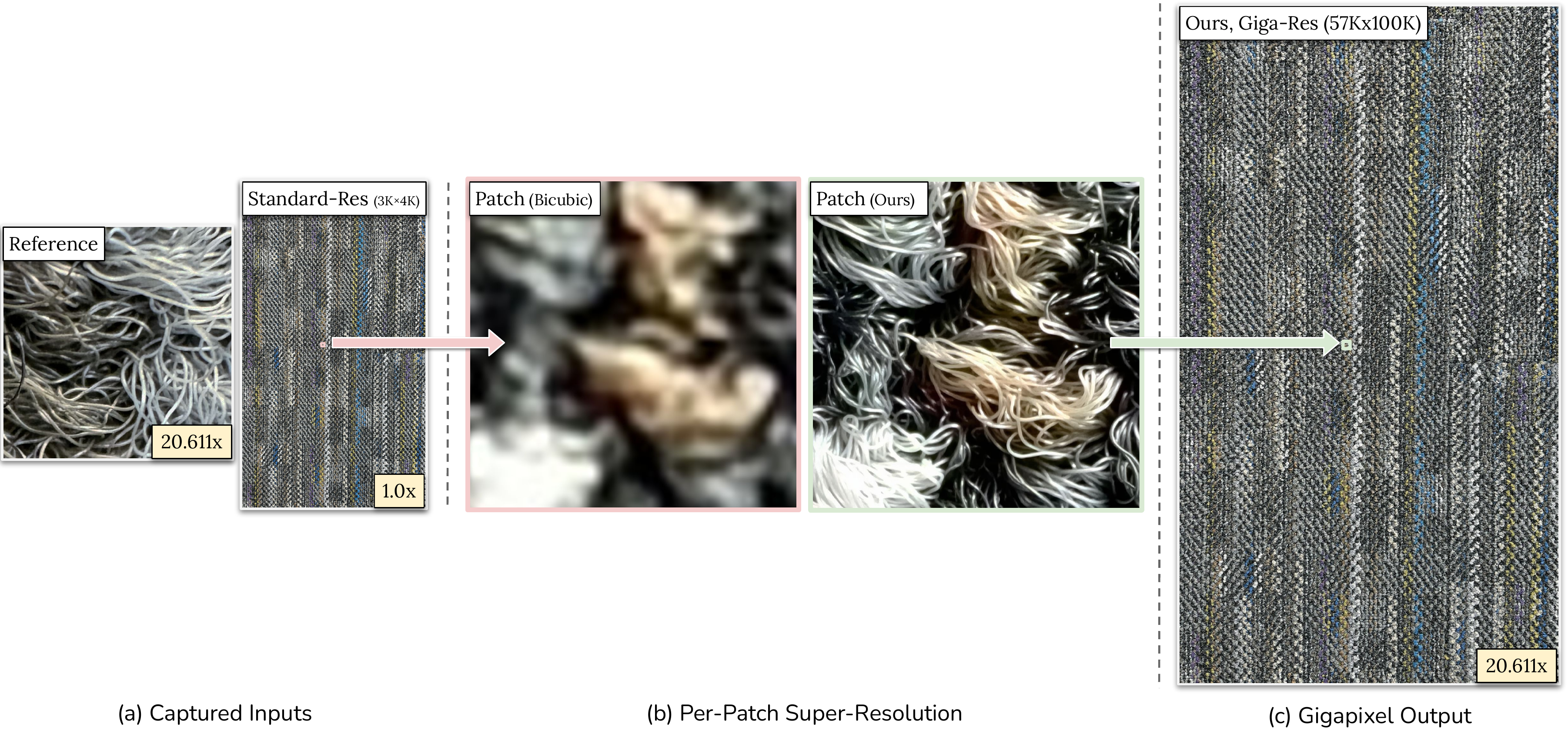}
    \caption{
    Given a standard-resolution image and a close-up reference of an object, our method faithfully restores object-specific details to each patch and merges the patch-wise predictions into a coherent gigapixel-resolution image (20x zoom example shown above).
    Visit our project page at \textcolor{lightblue}{\href{https://ultra-zoom.github.io}{ultra-zoom.github.io}} to interactively view all our full-resolution results, ranging from 6x to 30x in scale, and 0.25 to 5 gigapixels in output resolution.
    }
    \label{fig:teaser}
\end{teaserfigure}

\maketitle
\section{Introduction}
\label{sec:intro}

When photographing an object, we often want both a full overview and fine-grained material details. However, everyday cameras have fixed resolution, forcing a trade-off between coverage and detail. For example, product images typically show a low-detail full view alongside isolated close-ups, which limits the ability to examine the entire product in high resolution. Gigapixel images allow one to navigate the entire image in high detail, but creating such images typically requires specialized systems that capture and stitch together thousands of high-resolution photos \cite{KUDC07}.

While such exhaustive capture is necessary for complex scenes like city skylines, where each subregion contains unique structures and details, we argue that a single object is more self-similar and characterized by a limited set of material patterns. A small number of close-up snapshots often suffices to cover the range of textures. From these sparse examples, it is possible to infer the missing fine details across the entire object in a plausible and truthful manner.

We introduce \textit{UltraZoom}, a system that enables gigapixel-scale imaging of objects using only a phone camera and a casual, handheld capture process. Given a full image and one or more example close-ups, UltraZoom upscales the full image to match the scale and material details of the close-ups. The result is a unified, high-resolution image where local details are photorealistic and faithfully reflect the object's actual appearance.

To achieve this, we adapt a pretrained generative model by constructing a per-instance paired dataset from the close-ups, and learning how to map low-resolution patches to their high-resolution counterparts. During inference, we apply the model in a sliding-window fashion across the full image. A key challenge lies in constructing these training pairs, which requires registering close-ups within the full image despite significant scale differences and repetitive structures. We introduce a simple yet robust method for achieving such registration on casual captures of arbitrary materials.

Together, these components form a system that generates faithful, photorealistic gigapixel images from minimal input, enabling seamless pan-and-zoom exploration of an object in high detail.
Full-resolution results can be viewed interactively at \textcolor{lightblue}{\href{https://ultra-zoom.github.io}{ultra-zoom.github.io}}.


\vspace{0.5em}
\noindent Our contributions are:
\begin{itemize}[topsep=3pt,parsep=1ex,leftmargin=*]
\item A system for generating gigapixel-scale zoomable imagery using only sparse, casual handheld captures.
\item A simple, robust method for registering close-ups to a global image under unconstrained settings.
\end{itemize}

\begin{figure*}
    \centering
    \includegraphics[width=1\linewidth]{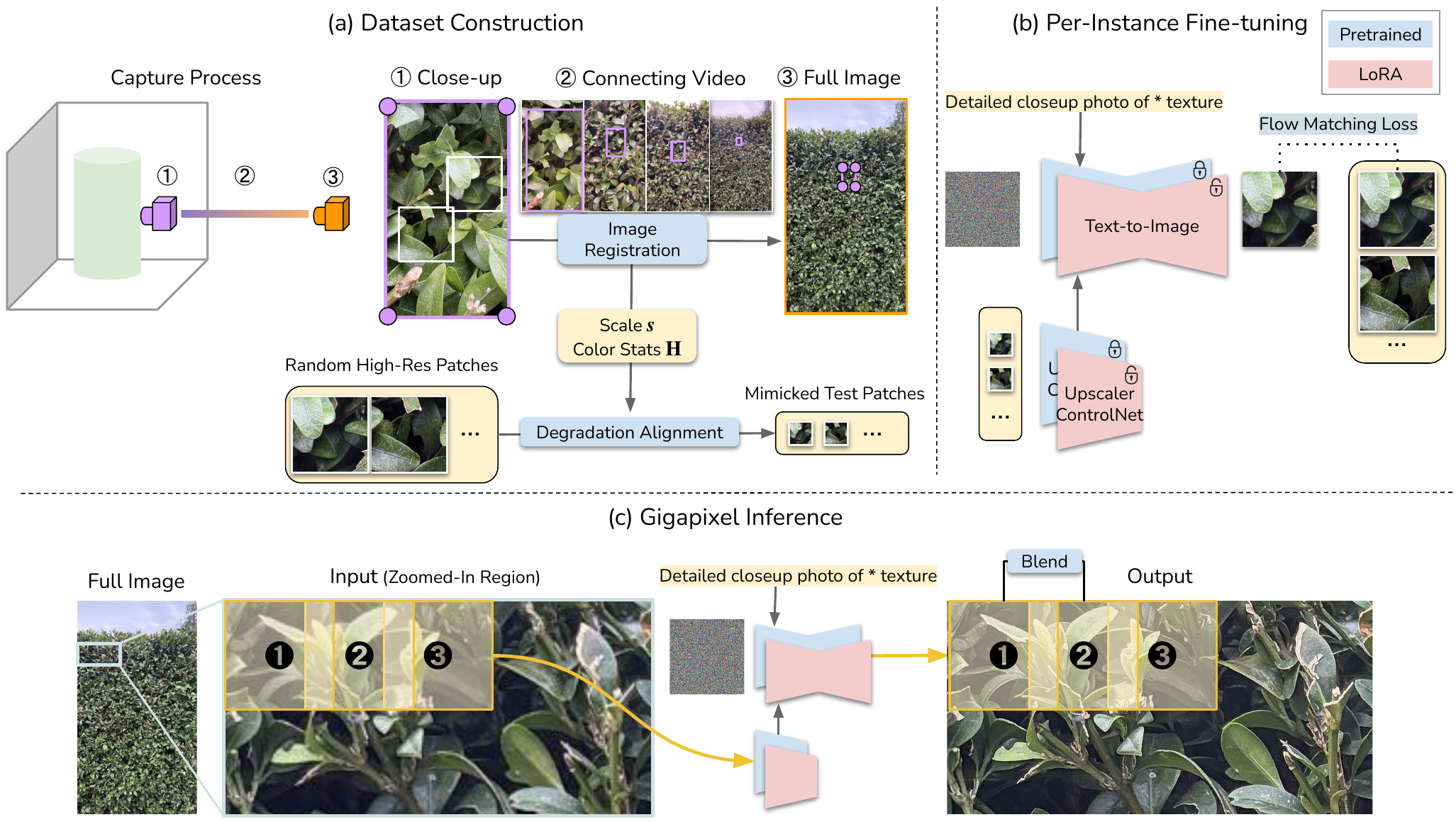}
    \caption{
    \textbf{Method Overview.} (a) \textit{Dataset Construction}: For each scene, we capture a close-up, a full-view image, and a bridging video that connects the two views. We track the close-up region across the video to register it within the full image, estimating the relative scale $s$ and the color statistics $H$ of the matched region. These are used to construct a dataset of paired high-resolution and degraded image patches, designed to align with inference-time test patches.
    (b) \textit{Per-Instance Fine-tuning}: Next, we fine-tune a pretrained generative model
    ~\cite{flux2024}
    on the per-instance dataset with DreamBooth ~\cite{ruiz2023dreamboothfinetuningtexttoimage} and LoRA ~\cite{hu2021loralowrankadaptationlarge}, allowing the model to adapt specifically to the captured object’s appearance. (c) \textit{Gigapixel inference}: At inference time, due to GPU memory constraints, we divide the full-view image into sliding windows of patches, apply super-resolution to each patch, and blend the overlapping regions in both latent and pixel space to produce a coherent gigapixel result.}
    \label{fig:method}
\end{figure*}

\section{Related Work}
\label{sec:related}

\subsection{Reference-Based Super-Resolution}

In Reference-Based Super-Resolution (RefSR), external high-resolution (HR) images are provided to guide the enhancement of a low-resolution (LR) input. RefSR encompasses a range of problem settings, each involving different types or configurations of reference images.

Early work in RefSR downsample generic high-resolution images to construct a database of representative HR–LR patch pairs. To improve generalization, small patches are extracted and often preprocessed to retain only structural or frequency components (e.g., by removing color). At test time, LR patches from the input are matched to this database using retrieval or sparse-coding techniques to construct the HR output~\cite{freeman2002example,1315043,5466111}.

More recent works leverage deep neural networks to incorporate reference images from semantically or structurally similar scenes. These methods focus on adaptively transferring high-resolution details from the reference to the corresponding regions of the low-resolution input. Many employ attention mechanisms to match and fuse features of the reference and input, often guided by hierarchical structure or semantic information~\cite{yang2020learningtexturetransformernetwork,zhang2019imagesuperresolutionneuraltexture,lu2021masasrmatchingaccelerationspatial,pesavento2021attentionbasedmultireferencelearningimage}.

Other works focus on settings where per-instance reference images are available, either of the same identity captured at different times or multiple captures of the same scene. For the former, \cite{varanka2024pfstorerpersonalizedfacerestoration} proposes a personalized face SR method using cross-attention to inject person-specific features from reference images. While full-face images contain consistent structure that suits attention-based transfer, our setting operates on highly local texture patches due to memory constraints, where the lack of structural regularity makes detail transfer more ambiguous. The latter category involves light field images ~\cite{zheng2018crossnetendtoendreferencebasedsuper}, or a dual-camera setting using wide-angle and telephoto image pairs~\cite{cai2019toward, Chen_2019_CVPR, Zhang_2019_CVPR, Wang_2021_ICCV, Xu_2023_CVPR}. These methods typically assume closely aligned image centers and small scale differences, enabling precise alignment between the LR input and HR reference using keypoints~\cite{lowe2004}, dense optical flow~\cite{fischer2015flownetlearningopticalflow}, warping neural network, or loss designs that tolerate slight misalignment. The aligned pairs are then used for supervised training. In contrast, our setup involves handheld captures with regular and macro lenses, where larger distortions and scale gaps make accurate alignment challenging. Due to these challenges, we instead construct pixel-aligned LR–HR patch pairs from the close-ups alone and train a model to learn direct local LR-to-HR mappings. The main challenge lies in degrading the HR patches to match their corresponding regions in the LR input (obtained via our coarse registration method), enabling the model to generalize effectively at test time.

\subsection{Texture Synthesis}
Texture synthesis aims to generate new pixels that match the appearance and statistical properties of an exemplar texture. Early non-parametric methods~\cite{790383, Efros01, kwatra:2003:SIGGRAPH} achieve this by copying best-matched patches from a reference image into target regions, ensuring local coherency but often struggling to preserve global structure consistency. Recent advances in deep learning have significantly expanded the capabilities of texture synthesis. CNN-based methods~\cite{gatys2015texturesynthesisusingconvolutional, ulyanov2016texturenetworksfeedforwardsynthesis, bergmann2017learningtexturemanifoldsperiodic}, GANs~\cite{zhu2018multimodalimagetoimagetranslation, xian2018texturegancontrollingdeepimage, shaham2019singanlearninggenerativemodel}, and diffusion-based approaches~\cite{wang2024infinitetexturetextguidedhigh} enable the generation of diverse and globally consistent textures. Our work also synthesizes new texture pixels from exemplars, but under the constraint of a low-resolution input that defines the global image structure, requiring the synthesized texture details to align with underlying content.

\subsection{Extreme-Scale Super-Resolution}

Extreme-scale super-resolution aims to reconstruct high-resolution outputs from significantly lower-resolution inputs, often beyond the typical 2–4× range. The large scale gap poses a major challenge: synthesizing high-frequency details not present in the input while maintaining consistency with the overall structure of the LR image.

One common approach is to progressively build up to the target scale with multi-scale designs \cite{shang2020perceptualextremesuperresolution} or cascaded models \cite{ho2021cascaded}. However, these methods operate at discrete scales and struggle to generalize beyond their training range. Cascading also introduces cumulative error and increases inference time.

Arbitrary-scale super-resolution addresses this by modeling the image in continuous space. Some methods condition explicitly on the desired scale \cite{chai2022anyresolution, hu2019metasrmagnificationarbitrarynetworksuperresolution}, while others reconstruct an implicit representation that allows querying at arbitrary coordinates \cite{Chen_2021_CVPR, becker2025thera, xu2022ultrasrspatialencodingmissing, peng2025pixel}. While these approaches offer flexibility, they often produce overly smooth results at extreme scales due to limited model capacity and difficulty recovering high-frequency details far beyond the training distribution.

Another line of work enables large-scale SR through per-instance or per-domain priors. Per-instance methods \cite{varanka2024pfstorerpersonalizedfacerestoration, ruiz2023dreamboothfinetuningtexttoimage} adapt the model to each input using fine-tuning or reference-guided attention to inject identity-specific features. Per-domain methods \cite{zhang2020texturehallucinationlargefactorpainting, sharma2024earthgengeneratingworldtopdown} leverage existing gigapixel-scale images (e.g., paintings, satellite imagery) to construct paired LR–HR data for supervised training. Our work is closely related but differs in that it does not assume access to pre-existing gigapixel images; instead, we construct the dataset from a regular-resolution image and a sparse set of close-up captures.



\begin{figure*}[t]
    \centering
    \includegraphics[width=1\linewidth]{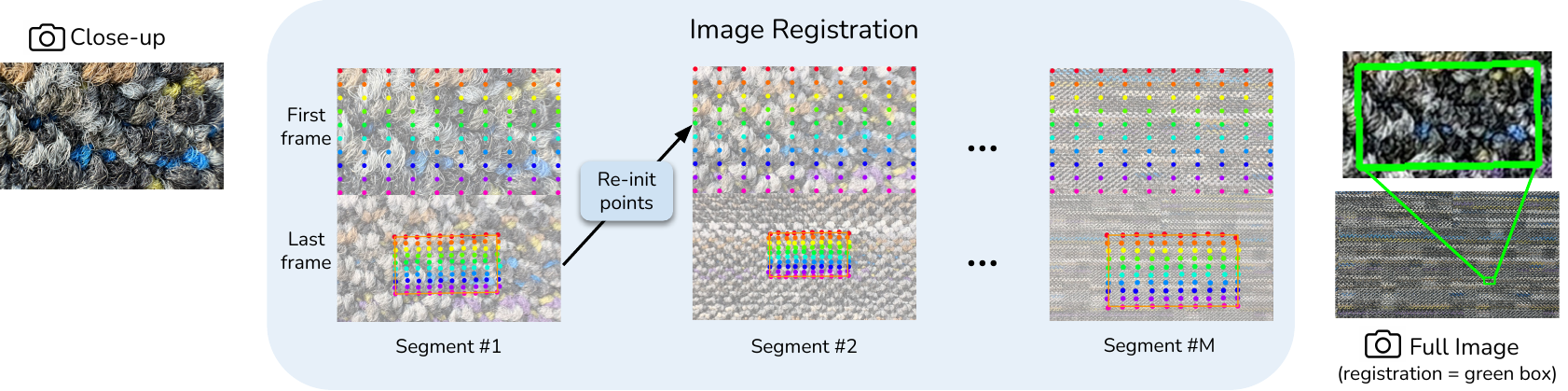}
    \caption{\textbf{Close-up-to-full registration.} Given a close-up, the full image, and a connecting video, we first split the video into shorter segments to improve point tracking, as the field of view changes rapidly. At the start of each segment, we initialize a grid of points and track them across frames within the segment. A 2D similarity transform is then estimated for each segment using RANSAC over the tracked points. These transforms are sequentially chained to produce the final transform that maps the close-up to the full image. The registration result is visualized in the green box (right).}
    \label{fig:tracking}
\end{figure*}

\section{Method}
\label{sec:method}

Given one or more example close-ups of an object and a regular-resolution image showing the entire object, our goal is to upscale the full-view image to the same scale as the close-ups, producing a gigapixel-resolution output with details faithful to the exemplars and structure aligned with the original full view.
Our system (illustrated in Fig.\ref{fig:method}) consists of three stages: dataset construction, per-instance fine-tuning, and gigapixel inference.

\subsection{Dataset Construction}

\textbf{Capture Process.} We use an iPhone with macro-lens mode for data collection. Given an object, we capture a minimal collection of $\mathcal{N}$ close-up images $\mathcal{C} = \{C_1, \dots, C_N\}$ to cover fine surface details, followed by a full-shot image $F$ that serves as the low-resolution input. To guide spatial registration, we also capture a sequence of short bridging videos $\mathcal{V} = [V_1, \dots, V_N]$, where each $V_i$ connects $C_i$ to $C_{i+1}$, and the final video $V_N$ connects $C_N$ to $F$. Camera orientation is kept consistent across all captures. In Fig.\ref{fig:method}a, we show a simplified case where only one close-up is captured.

\textbf{Motivation for Image Registration.} Given the captured images, our goal is to construct a paired dataset for supervised per-instance fine-tuning. However, the close-ups and full-view images are taken with different lenses and from different distances, resulting in large variations in perspective, color and noise levels, along with disocclusions and image center misalignment due to handheld capture. These factors make it extremely difficult to establish exact pixel-to-pixel correspondences between close-ups and their corresponding low-resolution pixels in the full image.

To address this, we construct pixel-aligned pairs using only the close-ups and apply degradation to simulate the appearance of the full-view image, enabling the model to generalize effectively during inference when applied to upscale the full image. This degradation process requires: (1) estimating the relative scale between close-ups and the full image for proper downscaling, and (2) identifying the corresponding region in the full image to match color statistics and other degradation characteristics. Both steps require coarse image registration, which we describe next.


\textbf{Image Registration.} Direct registration using SIFT or learning-based methods is challenging due to the highly repetitive nature of textures and significant scale differences between the close-ups $\mathcal{C}$ and the full view $F$. Instead, we leverage the bridging videos $\mathcal{V}$ and a state-of-the-art point tracking method ~\cite{karaev2024cotracker3} to track points across the full sequence and accumulate the transforms for final registration. Specifically, for each video $V_i$, we track a grid of 2D points and estimate a similarity transform $T_{V_i}$ between the first and the last frame using RANSAC \cite{fischler_bolles_1981}. By chaining these transforms across $V$, we obtain the cumulative transform from the close-up frame $\mathcal{C}_i$ to the full image $F$:

\begin{equation}
T_{\mathcal{C}_i \rightarrow F} = T_{V_N} \cdot (T_{V_{N-1}} \cdot (\cdots (T_{V_{i+1}} \cdot T_{V_{i}}) \cdots ))
\end{equation}

\noindent Lastly, we extract the scale $s$ from the upper-left $2 \times 2$ submatrix, which corresponds to the rotation and scaling component of the final transform  $T_{\mathcal{C}_i}$:
\begin{equation}
    s = \sqrt{\det\left( T_{\mathcal{C}_i \rightarrow F[:2, :2]} \right)}
\end{equation}

For $V_N$, the video dollying out from the last close-up $C_N$ to the full image $F$, the content in view changes quickly, so we divide the video into short segments $V_N = [S_1, \dots, S_M]$ and perform point tracking within each segment, allowing points to be reinitialized at the start of each segment for improved robustness. The per-segment transforms are chained similar to how per-video transforms are chained. We visualize this process in Fig.\ref{fig:tracking}.

\textbf{Degradation Alignment.} With registration complete, we minimize the train-test domain gap by degrading the high-detail close-ups to match the appearance of their corresponding regions in the lower-detail full image. We first address color inconsistency caused by varying white balance across captures. For each close-up $C_i$, we use the estimated transform $T_{\mathcal{C}_i \rightarrow F}$ to localize its corresponding region in the full image $F$, denoted $F_{C_i}$. We then extract its color statistics $H(F_{C_i})$ and apply color matching \cite{10.7717/peerj.453} to $C_i$ to obtain a color-corrected version $\tilde{C}_i$.

Next, we simulate degradation to match the appearance of the full image. We first apply bicubic downsampling to $\tilde{C}_i$ using the estimated scale factor $s$, followed by an additional 2x downsampling to mimic optical blur from distant capture. Since both the close-ups and the full image are JPEG-compressed on a standard iPhone, with artifacts appearing at different scales, we apply JPEG compression (quality = 75) when $F_{C_i}$ exhibits significantly more compression artifacts than the downscaled $\tilde{C}_i$. The fully degraded close-up, incorporating all degradation steps, is denoted as $D(\tilde{C}_i)$.



\subsection{Per-Instance Fine-tuning}
To minimize prior-based hallucination and ensure faithful reconstruction of object details, we fine-tune a pretrained generative model on the instance-specific dataset we construct from the close-ups. To fully leverage the pretrained model's high-fidelity generation capabilities, we freeze its weights and optimize only low-rank adaptations of the weight matrices, following~\cite{hu2021loralowrankadaptationlarge}.

As shown in Fig. \ref{fig:method}b, during training, we sample random patches $\mathbf{c} \sim \mathcal{P}(\tilde{C})$ and their corresponding degraded versions $\mathbf{d} \sim \mathcal{P}(D(\tilde{C}))$, where $\tilde{C} \in \{ \tilde{C}_1, \dots, \tilde{C}_N \}$ is drawn uniformly from all color-corrected close-ups, and $\mathcal{P}$ denotes a random patch extraction operator. The training objective is a flow matching loss:

\begin{equation}
\mathcal{L}_{\text{FM}} = \mathbb{E}_{\mathbf{c}_t, t, \mathbf{d}} \left[ \left\| \hat{\mathbf{u}}_\theta(\mathbf{c}_t, t, \mathbf{d}, \mathbf{y}) - \mathbf{u}(\mathbf{c}_t, t) \right\|_2^2 \right]    
\end{equation}

\noindent Here, $\mathbf{c}$ is the clean, high-detail patch from a color-corrected close-up, $\mathbf{d}$ is the corresponding degraded, lower-detail patch, and $\mathbf{c}_t$ is a noisy version of $\mathbf{c}$ at time step $t$. $\mathbf{u}$ is the target velocity from the forward diffusion process, and $\hat{\mathbf{u}}_\theta$ is the model's predicted velocity. $\mathbf{y}$ denotes the fixed per-instance text prompt used for conditioning.

\subsection{Gigapixel Inference}
Due to the gigapixel output size and GPU memory constraints, we perform inference on sliding local windows (see Fig.\ref{fig:method}c). Overlapping latent regions are averaged following~\cite{bartal2023multidiffusionfusingdiffusionpaths}, and overlapping RGB values are blended when stitching decoded pixels. However, noticeable boundary artifacts can still emerge (also observed in~\cite{wang2024}) due to patch discrepancies and repeated boundaries across steps, which cause neighboring patches to evolve inconsistently and diverge over time. To mitigate this, we introduce stride variation across denoising steps by gradually increasing the stride. This helps reduce artifact accumulation along fixed seams and lowers inference time by reducing the total number of sliding windows. See implementation details in supplemental.


\begin{figure*}
    \centering
    \includegraphics[width=1\linewidth]{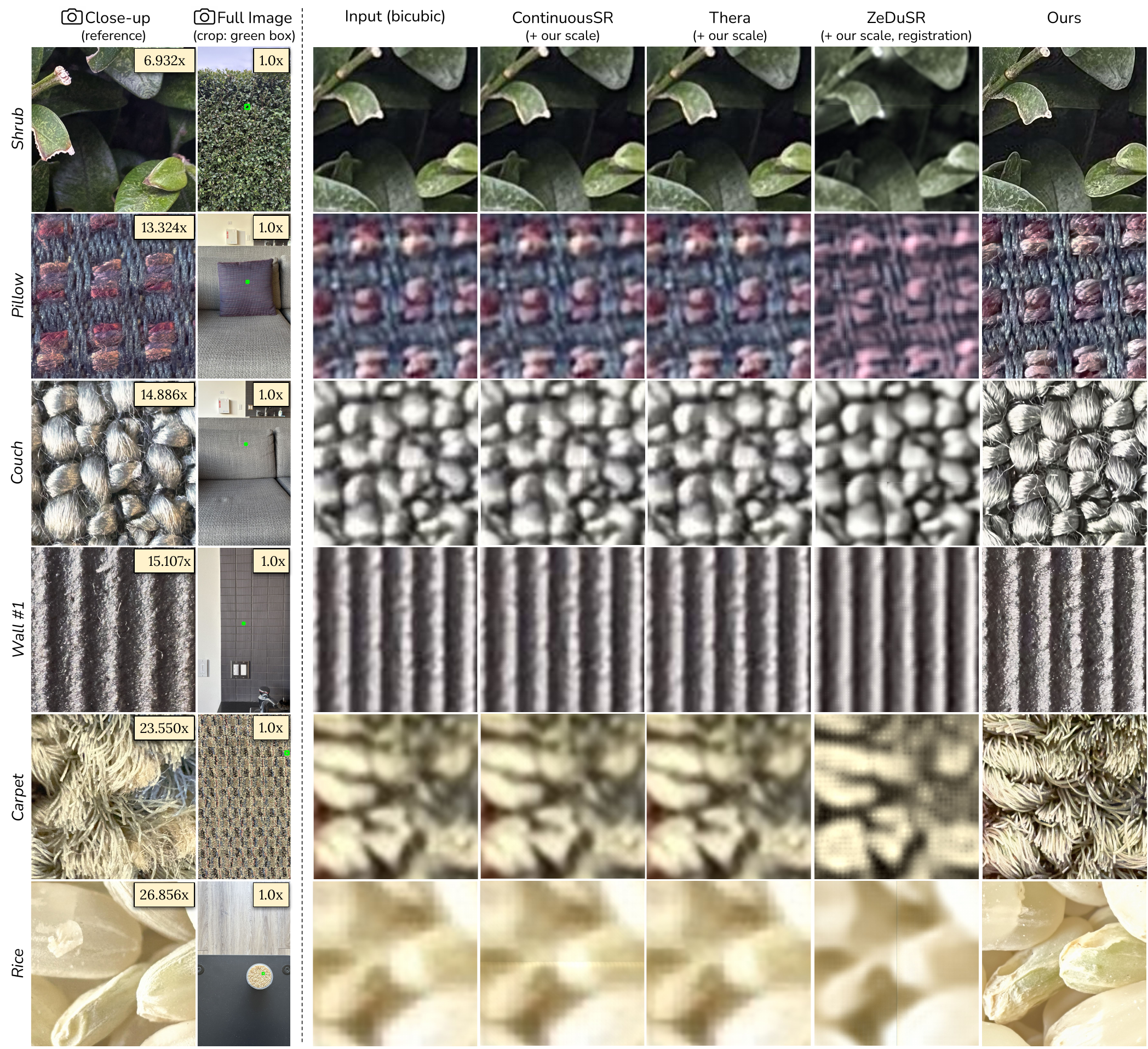}
    \caption{\textbf{Qualitative Comparison.} Rows are ordered from low to high scale. For each example, we compare $1024 \times 1024$ patches across methods. From left to right: a $1024 \times 1024$ crop from the captured close-up (reference), full image with patch location (green box), low-resolution input patch bicubic-upsampled to $1024 \times 1024$, three baselines, and our result. While the output patch may not be perfectly aligned with the reference, it is sampled near the reference, and the same crop is used across all methods for fair comparison. Qualitatively, our method achieves the best visual fidelity and consistency with the exemplar.
    }
    \label{fig:qualitative}
\end{figure*}

\section{Results}

We evaluate our method on 15 examples of everyday objects captured using an iPhone 16, with scale factors ranging from 6x to 30x. Among them, 2 examples contain multi-material objects and include multiple close-up captures to cover different surface materials. The following sections elaborate on baseline methods, quantitative and qualitative evaluations, ablations, and details of computational cost.

\subsection{Baseline Methods}

We compare with three baseline methods: (1) Thera ~\cite{becker2025thera}, (2) ContinuousSR ~\cite{peng2025pixel}, and (3) ZeDuSR ~\cite{Xu_2023_CVPR}.
Since Thera and ContinuousSR do not estimate scale and ZeDuSR relies on SIFT-based image registration (which fails on our data), we supply all methods with our estimated scale factors and, for ZeDuSR, our registration. Note that boundary blending is not implemented for any baseline, which may lead to visible seams in their outputs.

Thera and ContinuousSR are arbitrary-scale super-resolution methods based on implicit neural representations, allowing direct application of floating-point scale factors. Due to GPU memory constraints and their training resolution, we divide the 4K full-view image into overlapping $256 \times 256$ patches and stitch the outputs.

ZeDuSR is a dual-camera super-resolution method that performs per-instance training on wide-angle and telephoto image pairs. We adapt it to our setting by using the full image and close-up as the input pair. Since the implementation requires a power-of-two scale factor, we round up our estimated scale and downsample the input image to match the resulting scale difference. We replace ZeDuSR’s SIFT-based registration with our own and keep the rest of the pipeline unchanged. During inference, we use overlapping $1024 \times 1024$ patches, roughly following the input shape used in their tiling inference code.

\begin{table}
    \centering
    \caption{\textbf{Quantitative Comparison.} We report low-resolution mean absolute error (LR MAE) to measure consistency with the input, and Patch-FID/KID to evaluate perceptual similarity to the captured close-ups. Our method achieves the best Patch-FID and KID scores, indicating superior visual quality and texture fidelity, while maintaining competitive LR consistency. We also include user study results for human evaluation of visual quality and consistency. Our method ranks highest in both dimensions, being selected 96.08$\%$ of the time as the best in quality and in consistency 79.41$\%$ of the time.
}
    
    \begin{tabular}{c|c|c|c|c|c}
        Method & LR-MAE $\downarrow$ & P-FID $\downarrow$ & P-KID $\downarrow$ & \multicolumn{2}{c}{User Study, Top-1 $\uparrow$}   \\
        & & & & Qual. & Consis. \\

        
        \hline
       ContSR  & 0.038 & 310.371 & 0.350 & 0.33\% & 4.58\% \\
       Thera  & \textbf{0.007} & 327.872 & 0.391 & 1.63\% & 12.09\% \\
       ZeDuSR  & 0.051 & 348.874 & 0.404 & 1.961\% & 3.92\% \\
       Ours  & 0.040 & \textbf{134.986} & \textbf{0.130} & \textbf{96.08\%} & \textbf{79.41\%} \\
    \end{tabular}
    \label{tab:quantitative}
\end{table}

\subsection{Quantitative Comparison}
\label{sec:quantitative}

\textbf{Metrics.} We quantitatively evaluate performance using two types of metrics. (1) Mean Absolute Error (LR-MAE) between the low-resolution input and the bicubic-downsampled high-resolution output, which measures super-resolution consistency. (2) Patch-FID and KID. Since the close-ups and corresponding regions in the generated high-resolution output are not pixel-aligned, we compute Fréchet Inception Distance (FID) and Kernel Inception Distance (KID) on randomly sampled patches from the real and generated images, assuming they follow similar patch-level distributions. We sample approximately 3000 patches of size $299 \times 299$, keeping the crop positions fixed across methods for fair comparison.

\textbf{User Study.} We also conduct a user study: for each example in our evaluation set, participants are shown a $1024 \times 1024$ patch from the real captured close-up alongside a nearby patch from each method's output, with patch position fixed across methods. They are asked to select the best result along two dimensions: visual quality and consistency with the exemplar in terms of texture detail. For the examples with multiple close-ups, each close-up is treated as a separate instance, expanding 15 examples into 18. We collect responses from 17 participants, each evaluating all 18 examples across both dimensions, resulting in 612 total responses.

\textbf{Discussion.} As shown in Tab.~\ref{tab:quantitative}, the two per-instance methods ZeDuSR and Ours have higher low-resolution mean absolute error (LR-MAE): 0.051 and 0.040, compared to 0.038 and 0.007 for ContinuousSR and Thera. This is expected, as both methods generate more high-frequency details, which may deviate slightly from the low-resolution input due to the lack of explicit consistency constraints during training. In terms of generation quality, our method significantly outperforms all baselines in Patch-FID and KID (134.986 vs. \(\sim\!300\)), demonstrating strong alignment with the quality and texture details of the captured close-ups. The user study further validates these results: our method is overwhelmingly preferred in visual quality (96.08$\%$) and achieves strong performance in consistency (79.41$\%$), highlighting its effectiveness while indicating potential for further improvement in suppressing subtle hallucinations and maintaining detail consistency with the exemplar.

\begin{figure*}
    \centering
    \includegraphics[width=1\linewidth]{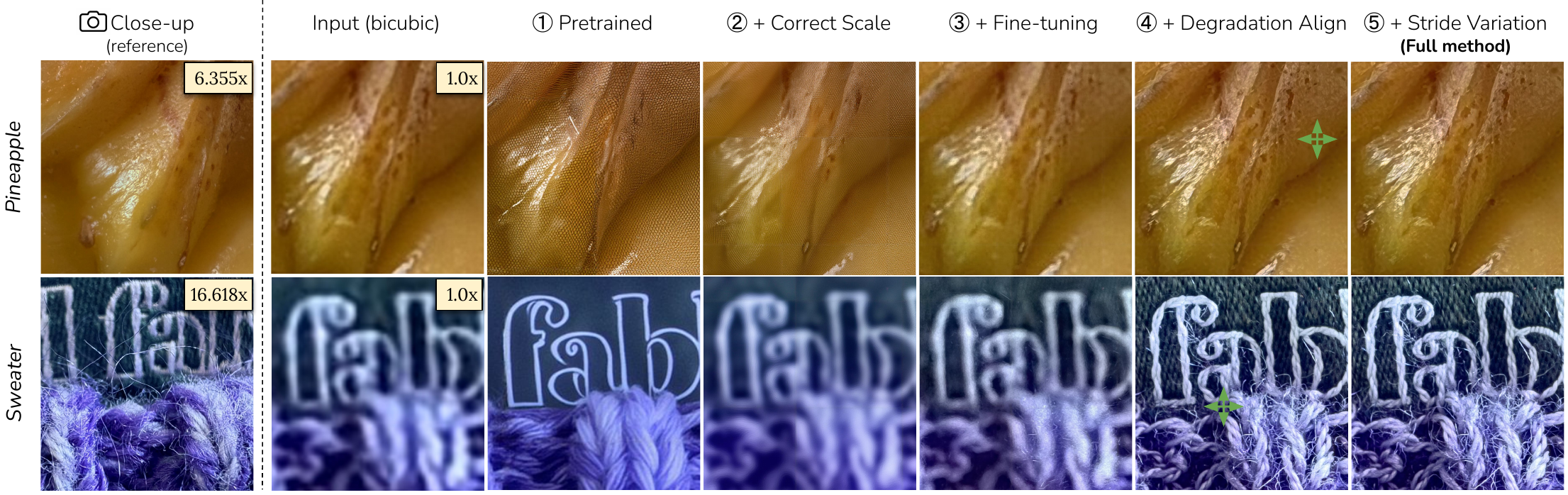}
    \caption{\textbf{Ablations.} We show the effect of individual method components on two examples. (1) Pretrained model at default 4x scale produces results with hallucinated details. (2) Using the correct scale without the close-up reference still results in hallucination and introduces artifacts due to poor generalization beyond the training scales. (3) Adding fine-tuning without proper degradation introduces instance-specific details, but the output suffers from domain mismatch. (4) Including degradation alignment reduces train-test gap and leads to improved results, but visible seams remain due to patch-wise inference without stride variation (see faint lines near green arrows). (5) Our full method resolves all issues, producing seamless and faithful gigapixel outputs.
}
    \label{fig:ablations}
\end{figure*}

\subsection{Qualitative Comparison}

We present qualitative comparisons across a range of scales in Fig.~\ref{fig:qualitative}. All baselines struggle to produce high-quality, detail-consistent results. Thera and ContinuousSR are trained on 2-4x discrete and 4-8x continuous scales, respectively. Despite the \textit{shrub} example having a 6.93× scale, which is within ContinuousSR’s range, it still fails to generate proper details, likely because the model was trained on natural images with standard fields of view, whereas our inputs are narrow FoV local patches, which fall outside its training distribution.

ZeDuSR also fails despite per-instance training. This may stem from its LR–HR alignment assumption: it warps the low-resolution, wide FoV image to be pixel-aligned with the high-resolution, narrow FoV image. However, our data cannot be easily aligned, resulting in training on misaligned LR–HR pairs, which may hinder learning. Additionally, ZeDuSR trains a lightweight model from scratch, whereas we fine-tune a high-capacity pretrained generative model. While their approach is well-suited for dual-camera setups with modest scale differences and minimal misalignment, our method is better equipped for recovering fine-grained details in extreme close-ups and handling casually captured inputs.

\subsection{Ablations}

We visualize the effect of each component of our method in Fig.~\ref{fig:ablations} with two examples. The most naive baseline applies the pretrained model at its original scale (4x), which is equivalent to standard single-image super-resolution. Without leveraging the close-up reference, the model may produce high-quality outputs but hallucinates details. When supplied with the correct scale, the pretrained model continues to hallucinate and struggles to generalize to scales beyond its training scale,  introducing artifacts (\textit{pineapple}) or failing to enhance the input meaningfully (\textit{sweater}).

Adding per-instance fine-tuning without proper degradation (i.e., without color matching, blurring, or JPEG compression) begins to introduce instance-specific details, but these details fail to integrate seamlessly with the structure of the low-resolution input, as the model is not trained to handle the appearance of the test-time degraded input. Including degradation alignment resolves this gap, but without stride variation, visible seams remain at inference window boundaries (faint lines extending from the marked green arrows). The final column shows our full method, which effectively addresses all these issues and produces seamless, high-fidelity results.



\subsection{Computational Cost}

All experiments are conducted on a single A100 GPU. Each training epoch takes approximately 22 minutes while inference time depends on the output resolution. For example, generating a $18672 \times 18672$ output with 28 denoising steps and stride variation results in 21,366 forward passes in total (763.07 forward runs per step). At 0.64 seconds per run, the total inference time is approximately 3.82 hours.

\section{Discussion and Limitations}
\label{sec:discussion}

We present a system for generating high-quality, faithful gigapixel images from a regular-resolution image and a set of close-ups. While our results demonstrate strong potential, several limitations remain that point to valuable directions for future work:

\paragraph{Slow Inference.}
Our current inference pipeline involves sliding-window generation with overlapping patches, which becomes computationally expensive at gigapixel scales. One promising direction is retrieval-based inference: for frequently occurring or similar patches, future work could cache or retrieve previously generated outputs instead of re-running the model, trading off some consistency for improved efficiency.

\paragraph{Per-Instance Fine-Tuning.}
To maximize generation quality, we perform per-instance fine-tuning by adapting a small set of model parameters to each object. While this yields strong results, it requires retraining for every new object, which is inefficient. An alternative would be to develop a general model that directly takes in the LR image and a few reference patches, and propagates high-frequency details across the object without fine-tuning.

\paragraph{Degradation Alignment.}
Our method relies on degrading the close-up patches to mimic the characteristics of patches from the full image, due to the difficulty of registering macro and regular photos. In this work, we manually choose the degradation operations (e.g., blur, downsampling, JPEG artifacts) to simulate the LR appearance. In future work, it would be valuable to automate this process by optimizing degradation parameters to best match the patch distributions between HR and LR views, potentially using a learned degradation model or domain adaptation framework.


\paragraph{Lack of Global Context.}
Currently, the model operates on local image crops without access to global information, such as the position of the patch within the full image or the broader structural context of the object. This limits the model’s ability to reason about object-level consistency or to apply the correct high-resolution details in ambiguous regions. Injecting global context through positional encoding, spatial layout features, or hierarchical features could help the model make more coherent and informed predictions, particularly when synthesizing large-scale images with long-range dependencies.

\clearpage
{
    \small
    \bibliographystyle{ACM-Reference-Format}
    \bibliography{main}
}

\clearpage
\begin{figure*}
    \centering
    \includegraphics[width=1\linewidth]{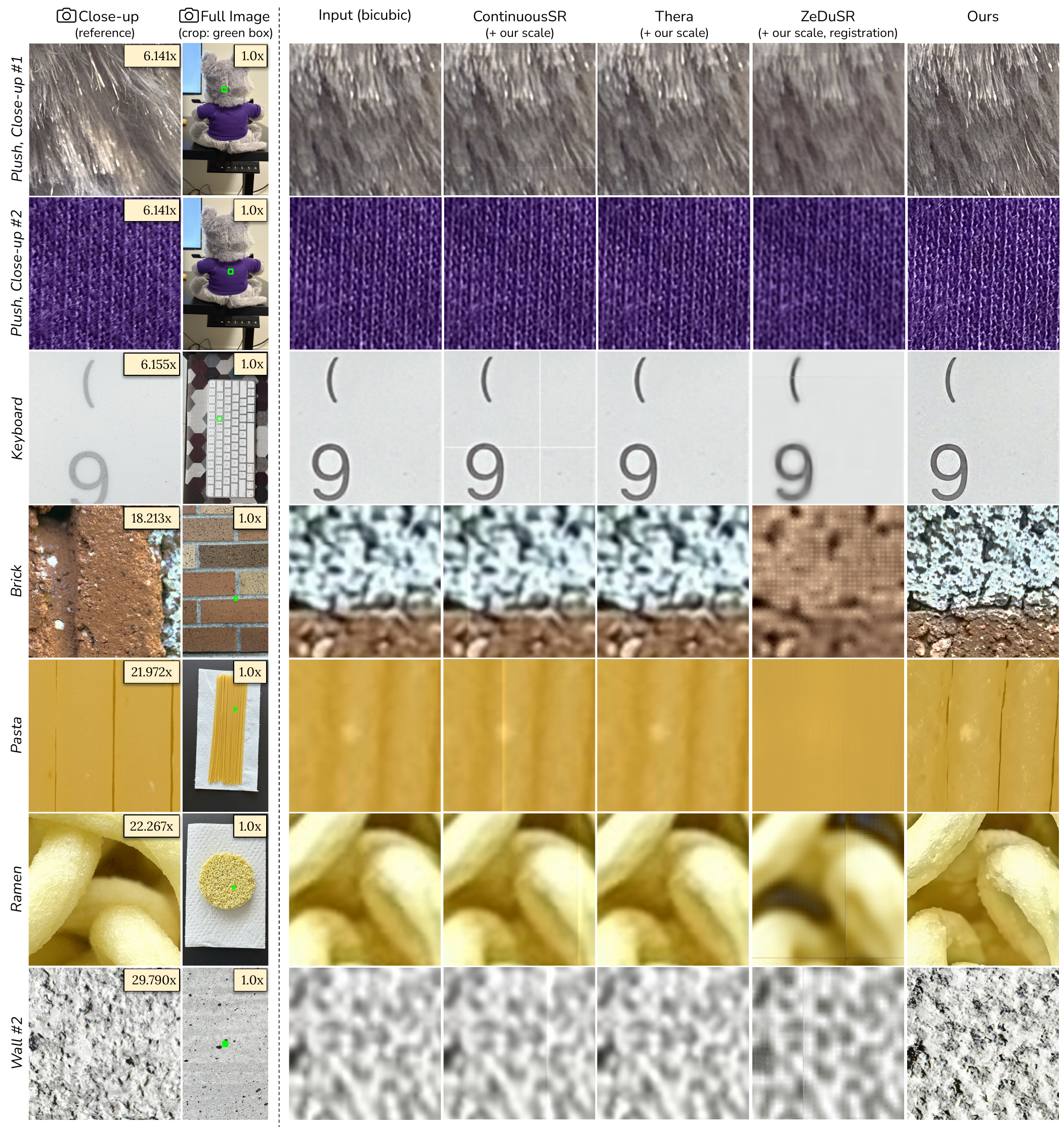}
        \vspace{-1em}
        \caption{\textbf{Additional Results.} 
        We present further qualitative comparisons between baseline methods and ours on additional captured examples. These results continue to demonstrate the strong visual quality and exemplar consistency achieved by our approach.
        Visit the supplemental video or \textcolor{lightblue}{\href{https://capybara1234.s3.amazonaws.com/demo_page/index.html?ts=12345}{the demo webpage}} to see the full-resolution results in an interactive interface.
    }
    \label{fig:qualitative}
\end{figure*}

\end{document}